\PassOptionsToPackage{pdftex,dvipsnames}{xcolor}
\documentclass[letterpaper, 10 pt, conference]{ieeeconf}  

\IEEEoverridecommandlockouts                             
\usepackage{graphics}
\usepackage{epsfig} 
\usepackage{mathptmx} 
\usepackage{times} 
\usepackage{amsmath} 
\usepackage{amssymb}  
\usepackage{subcaption}
\usepackage{graphicx}
\usepackage{float}
\usepackage{bm}
\usepackage{caption}
\usepackage{mathtools}
\usepackage{stfloats}
\usepackage{mathrsfs}
\usepackage{multirow}
\usepackage{makecell}
\usepackage{vcell}
\usepackage{booktabs}
\usepackage{diagbox}
\usepackage{csquotes}
\usepackage{cite}
\usepackage{tikz}
\usepackage{lipsum}
\usepackage{schemata}

\usepackage[hyphens]{url}
\usepackage[hidelinks]{hyperref}
\hypersetup{breaklinks=true}
\usepackage{cite}

\usepackage{algorithm}
\usepackage[noend]{algpseudocode}

\algrenewcommand\algorithmicforall{\textbf{for each}}
\algrenewcommand\algorithmicindent{.8em}

\algdef{SE}[DOWHILE]{Do}{doWhile}{\algorithmicdo}[1]{\algorithmicwhile\ #1}%
\algnewcommand\algorithmicforeach{\textbf{for each}}
\algdef{S}[FOR]{ForEach}[1]{\algorithmicforeach\ #1\ \algorithmicdo}

\usetikzlibrary{decorations.pathreplacing,calc}
\newcommand{\tikzmark}[1]{\tikz[overlay,remember picture] \node (#1) {};}

\newcommand*{\AddNote}[4]{%
    \begin{tikzpicture}[overlay, remember picture]
        \draw [decoration={brace,amplitude=0.5em},decorate,ultra thick,black]
            ($(#3)!(#1.north)!($(#3)-(0,3)$)$) --  
            ($(#3)!(#2.south)!($(#3)-(0,3)$)$)
                node [align=center, text width=2.5cm,  anchor=west,  rotate=90, pos=0.5, anchor=north, shift={(0,-0.3)}] {#4};
    \end{tikzpicture}
}%
\newcommand*{\AddNot}[4]{%
    \begin{tikzpicture}[overlay, remember picture]
        \draw [decoration={brace,amplitude=0.5em},decorate,ultra thick,black]
            ($(#3)!(#1.north)!($(#3)-(0,3)$)$) --  
            ($(#3)!(#2.south)!($(#3)-(0,3)$)$)
                node [align=center, text width=2.3cm,  anchor=west, rotate=90, pos=0.5, anchor=north, shift={(0,-0.3)}] {#4};
    \end{tikzpicture}
}%
\usepackage{hyperref}
\hypersetup{
colorlinks=true,
linkcolor=blue,
filecolor=magenta,
urlcolor=blue,
}

\usepackage[font=footnotesize]{caption}

\title{\LARGE \bf
UPPLIED: UAV Path Planning for Inspection through Demonstration 
}
\author{Shyam Sundar Kannan$^{1}$, Vishnunandan L. N. Venkatesh$^{1}$, Revanth Krishna Senthilkumaran$^{2}$, \\ and Byung-Cheol Min$^{1}$
\thanks{$^{1}$Shyam Sundar Kannan, Vishnunandan L. N. Venkatesh, and Byung-Cheol Min are with the SMART Lab, Department of Computer and Information Technology, Purdue University, West Lafayette, IN 47907, USA
{\tt\small \{kannan9,lvenkate,minb\}@purdue.edu}}%
\thanks{$^{2}$Revanth Krishna Senthilkumaran is with the SMART Lab, Department of Computer and Information Technology, Purdue University, and with the School of Electrical and Computer Engineering, Purdue University, West Lafayette, IN 47907, USA
{\tt\small senthilr@purdue.edu}}%
}

\begin{document}
\maketitle
\thispagestyle{empty}
\pagestyle{empty}

\begin{abstract}
In this paper, a new demonstration-based path-planning framework for the visual inspection of large structures using UAVs is proposed. We introduce UPPLIED: UAV Path PLanning for InspEction through Demonstration, which utilizes a demonstrated trajectory to generate a new trajectory to inspect other structures of the same kind. The demonstrated trajectory can inspect specific regions of the structure and the new trajectory generated by UPPLIED inspects similar regions in the other structure. The proposed method generates inspection points from the demonstrated trajectory and uses standardization to translate those inspection points to inspect the new structure. Finally, the position of these inspection points is optimized to refine their view. Numerous experiments were conducted with various structures and the proposed framework was able to generate inspection trajectories of various kinds for different structures based on the demonstration. The trajectories generated match with the demonstrated trajectory in geometry and at the same time inspect the regions inspected by the demonstration trajectory with minimum deviation. The experimental video of the work can be found at \href{https://youtu.be/YqPx-cLkv04}{https://youtu.be/YqPx-cLkv04}.
\end{abstract}

\section{Introduction}
\label{sec:intro}
Unmanned Aerial Vehicles (UAVs) have been increasingly employed in a variety of applications in recent years, such as surveying \cite{albani2017field}, agriculture\cite{anthony2014crop}, search and rescue \cite{xiao2017uav}, and package delivery \cite{kannan2022autonomous}. One promising area for the growth of UAVs is the vision-based structural inspection of large 3D structures, as they can significantly reduce inspection time and cost \cite{jordan2018state}. Large structures like buildings, airplanes, wind turbines, and more require regular visual inspection to identify defects such as cracks and dents on their surfaces \cite{metni2007uav}. To perform these vision-based inspections, UAVs must navigate around the structure and record visual information on its surface. Typically, UAVs are teleoperated or flown autonomously around the structure to conduct inspections. 

When UAVs are teleoperated for inspection, a task expert is usually required to pilot the UAV around the structure. Manual teleoperation enables great customizability in terms of the regions of the structure to be inspected and allows for adjustments to the flight trajectory to focus on critical regions. However, hiring an expert to pilot the UAV can be expensive, and inspecting large structures can be a tedious process \cite{vaughn_college_2022}. To overcome these limitations, several methods involving autonomous UAVs for inspection have been proposed \cite{almadhoun2018coverage, bircher2015structural, jing2020multi, jenssen2018automatic}.

\begin{figure}[t]
\centering
    \includegraphics[width=0.875\linewidth]{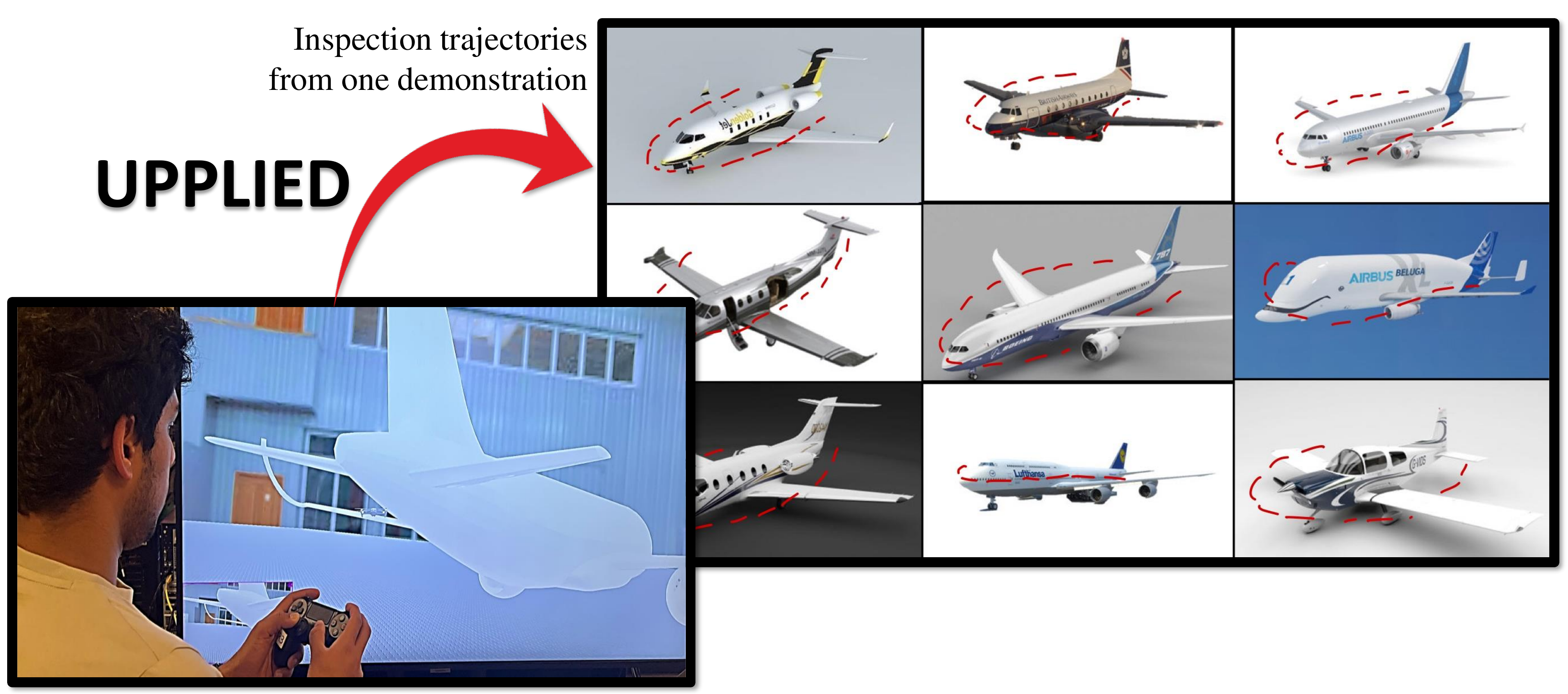}
\caption{An expert performing an inspection on the Airbus A320 aircraft, which is used as a demonstration model. The proposed UPPLIED leverages this demonstration to generate a trajectory for a UAV that can inspect similar regions to that of the demonstration trajectory on multiple different aircrafts that are similar in structure.}  
\label{fig:intro_pic}
\vspace{-7mm}
\end{figure}

Autonomous methods for UAV inspection typically involve computing paths for full coverage of the structure \cite{almadhoun2018coverage, bircher2015structural} or focusing on one specific type of structure \cite{jenssen2018automatic}. While these methods are more accessible, they may not always be appropriate. Complete coverage of the full structure may not always be necessary, and inspecting only certain critical regions of the structure may suffice. For example, during airplane inspections before a flight, only specific components like external sensors, engines, and landing gear are inspected \cite{airplane_safety_2017}. Therefore, complete coverage path planning methods are not well-suited for such scenarios, and customized planners to inspect specific regions of the structure are more efficient. However, generating customized paths to inspect specific regions can be challenging, and input from a task expert can be advantageous. 

In this work, we leverage the concept of learning from demonstration (LfD)  to develop a framework that learns from an expert's demonstration flight trajectory and generates a similar trajectory for structures of the same kind. For example, based on the human expert's trajectory to inspect certain regions of an airplane, we aim to generate a trajectory to inspect a similar region of another airplane, given that the structures are identical. As shown in Fig. \ref{fig:intro_pic}, a human demonstrator inspects the fuselage of an Airbus A320 airplane and the trajectory is recorded. Now, our goal is to generate a similar trajectory for inspecting the fuselage of another airplane. This approach is intended to bridge the gap between human teleoperated inspections and completely autonomous inspection by learning from a human demonstration and generalizing it so that it can be applied at scale. 

We propose UPPLIED: \textbf{U}AV \textbf{P}ath \textbf{PL}anning for \textbf{I}nsp\textbf{E}ction through \textbf{D}emonstration. The proposed framework estimates a new trajectory for inspecting one structure, given two structures of a similar kind (one for demonstration and the other as a target to generate an inspection trajectory), and a demonstration trajectory for inspecting one of the structures. In Fig. \ref{fig:intro_pic}, we show the demonstration trajectory from an expert for inspecting the fuselage of an Airbus A320 airplane which is then used by our framework to inspect the fuselage of multiple airplanes using the trajectory from one demonstration as a reference. To the best of our knowledge, this is the first work in the literature that employs UAVs for structural inspection where the inspection trajectory is generalized from expert demonstrations. It demonstrates a practical advantage in dealing with customized paths to inspect specific regions efficiently. 

The main contributions of our work are the following:
\begin{itemize}
    \item We introduce a framework where UAVs can learn to inspect a structure by leveraging the demonstration from an inspection expert which is used to generate trajectories for new structures of the same kind. 
    \item We extensively validate the proposed framework on numerous structures. We evaluate the similarity between the demonstrated and target trajectories, and compare the visibility between the two paths for all the structures.   
\end{itemize}




\section{Related Works}
\label{sec:rel_work}

\subsection{UAV for Inspection}
\label{sec:rel_uav_ins}

The use of Unmanned Aerial Vehicles (UAVs) for inspection is a well-studied topic, with extensive research on the inspection of structures such as power lines \cite{liu20163d}, bridges \cite{ ellenberg2016bridge}, buildings \cite{sankarasrinivasan2015health} and wind turbines \cite{mathe2015vision}. However, these methods are often specific to a single type of structure and not scalable to different types of structures. 

Coverage Path Planning (CPP) generates paths that enable the complete inspection of a structure. CPP has been used for the generalized inspection of various structures, where it uses the model of the structure to generate the path. Sampling-based CPP methods generate redundant viewpoints around the structure and optimize these viewpoints to achieve complete coverage \cite{bircher2015structural, almadhoun2018coverage, 9523743}. Multi-UAV-based CPP methods have also been developed to reduce the inspection time and cover larger areas \cite{kannan2019multi, jing2020multi}. However, the goal of all CPP methods proposed in the literature is to achieve complete coverage, which may not be desired in all scenarios. In our approach, we focus on generating trajectories that cover specific regions of the structure that are the focus of inspection, rather than aiming for complete coverage.

\subsection{Learning from Demonstration}
\label{sec:rel_lfd}
In the context of Learning from Demonstration for path planning or learning trajectories, there are several learning methods \cite{xie2020robot}. The common methods used in literature can be classified into two main approaches - Imitation Learning (IL) and Inverse Reinforcement Learning (IRL). In IL-based approaches, the robot attempts to imitate the demonstrated trajectory by fitting the demonstration data, which consists of several collected trajectories \cite{fang2019survey}. IL methods require a significant number of demonstrations depending on the complexity of the task \cite{pastor2009learning}. To this end, one-shot imitation methods that require just one demonstration have been proposed \cite{duan2017one}, but they still require numerous demonstrations for a base task. Furthermore, these IL methods attempt to imitate the demonstrated trajectory data but are oblivious to high-level features (such as state changes, key points, etc.) that influence the trajectory \cite{asfour2008imitation}. In the case of IRL approaches, they learn the parameterized rewards that can shape a policy \cite{arora2021survey}. This enables such approaches to take real-time decisions for path planning given an environment state and a set of actions. IRL approaches generalize well to new situations \cite{boularias2011relative}, but they pose the caveat of having to rely on a large set of demonstrations, as it helps shape the rewards, and they are also computationally expensive. 

When considering the inspection task as a real-time application, obtaining a large number of demonstrations becomes a tedious and expensive process, since it involves hiring an expert and inspecting multiple large structures. To this end, we prioritized an approach that uses a single demonstration to learn the inspection task (single-shot learning). Our approach encodes the demonstrated trajectory into segments consisting of high-level inspection points. These high-level inspection points act as a guide to generalize the trajectory from the demonstration to new and unseen models.  

\section{Problem Description}
\label{sec:problem}
This work addresses the problem of vision-based structural inspection path planning, which involves generating a trajectory for the complete or partial inspection of a 3D structure based on a demonstration path and the 3D CAD model of the structure. The demonstration path is obtained from an expert who teleoperates the UAV during structure inspection. 

In this work, we consider 3D models of two structures: a demonstration model ($\mathbb{M}_D$) used for expert demonstration and a target model ($\mathbb{M}_T$) to be inspected, that are aligned in the same orientation. Although $\mathbb{M}_D$ and $\mathbb{M}_T$ are structures of the same kind, they differ in their geometry, scale, and other finite features. Let $\mathbb{P}_D$ be the path trajectory that inspects the structure completely or partially, covering a few parts of the structure with Model $\mathbb{M}_D$. We assume that $\mathbb{P}_D$ is optimal and noise-free and is obtained from an expert. For a given target model $\mathbb{M}_T$, a target path trajectory $\mathbb{P}_T$ has to be generated that optimizes for the structural similarities between $\mathbb{M}_D$ and $\mathbb{M}_T$ and generalizes from the demonstrated trajectory $\mathbb{P}_D$. The goal is to generate a $\mathbb{P}_T$ that can efficiently inspect the target model $\mathbb{M}_T$ by taking cues from the demonstration model and path.  

 \begin{figure*}
  \centering
  \includegraphics[width=1\textwidth]{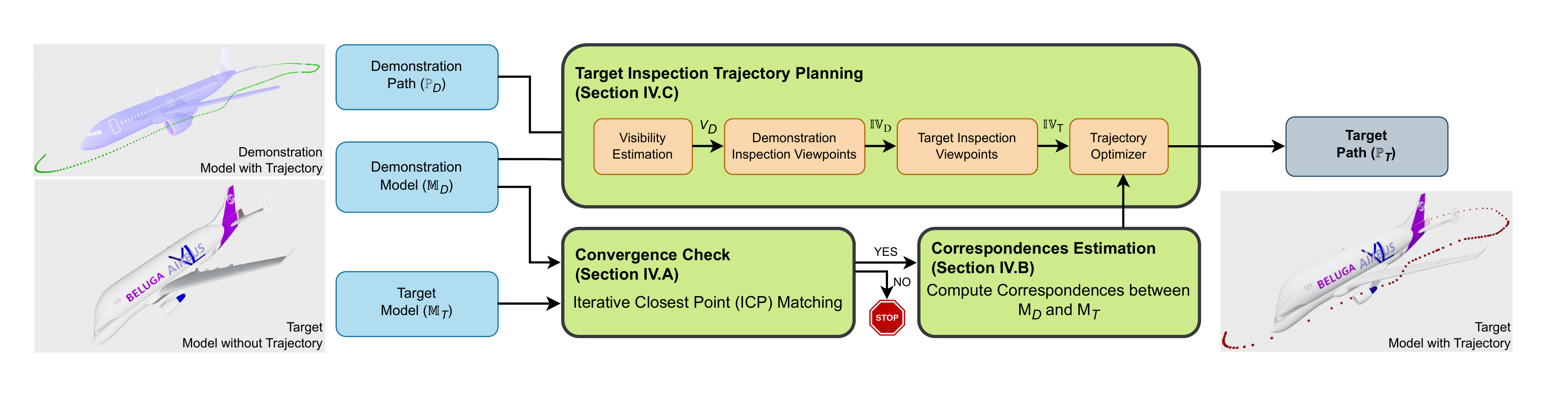} 
  \vspace{-20pt}
    \caption{A schematic overview of the proposed path-planning framework for structural inspection using UAVs. The inputs to the framework are the demonstration path (\textcolor{green}{green}) and models of the demonstration and the target structures. The demonstration model and target model are fed to the convergence checker (Sec. \ref{sec:conv_check}) which uses iterative closest point to check for convergence. If convergence is successful, the correspondences between both the input models are estimated (Sec. \ref{sec:corr_est}). The target inspection trajectory planning  (Sec. \ref{sec:traj_comp}) takes as inputs the demonstration model and demonstration path. Using these inputs, visibility is estimated from which the inspection viewpoints are computed. The inspection viewpoints are then optimized using the correspondences obtained from (Sec. \ref{sec:corr_est}) resulting in the target path (\textcolor{Maroon}{red}) for inspection. }
   \label{fig:overall_system}
   \vspace{-12pt}
\end{figure*}

\section{Methodology}
\label{sec:methodology}
In this paper, we propose a novel path-planning framework for structural inspection using UAVs. The framework learns from a demonstrated path used to inspect a structure and generates a new path for the inspection of a similar structure. The proposed framework can be summarized as follows: first, a similarity sanity check is performed to ensure that the two models are similar before proceeding with further steps. Next, the correspondences between the two models that associate the similar regions between them are computed. Then, the demonstrated path is used to estimate a coarse path for inspecting the new structure. Finally, this estimated path is optimized using the Gauss-Newton algorithm to generate the final inspection path. The overall structure of the proposed path-planning framework is shown in Fig. \ref{fig:overall_system}.

\subsection{Convergence Check}
\label{sec:conv_check}
The two models, $\mathbb{M}_D$ and $\mathbb{M}_T$, must be similar to a certain extent so that the trajectory demonstrated on $\mathbb{M}_D$ can be used to generate a trajectory to inspect $\mathbb{M}_T$. Let $\Psi_D$ and $\Psi_T$ be the point clouds that represent the models $\mathbb{M}_D$ and $\mathbb{M}_T$, respectively. The two-point clouds are sub-sampled to a uniform distribution using a Voxel Grid Filter \cite{Rusu_ICRA2011_PCL} to obtain sub-sampled point clouds of the structure's surface $\kappa_D$ and $\kappa_T$, respectively, which represent the high-level geometry of the structures. Then, $\kappa_D$ and $\kappa_T$ are converted into equivalent scales, such that their bounding boxes are identical. Scaling is performed non-uniformly across the three axes to convert the point clouds to equivalent scales. Now that the sub-sampled point clouds have similar scales, they are aligned with each other for a fixed number of iterations using Iterative Closest Point (ICP) Matching. The convergence fitness score \cite{Rusu_ICRA2011_PCL} from the ICP is used to evaluate the similarity. If the fitness score is too high, it implies that the models are not similar. In the implementation, it is checked if the fitness score is lesser than a threshold, $\gamma$. If the fitness score is greater than the threshold, a new trajectory is not estimated. In Algorithm \ref{alg:corr_alg}, lines 1 to 6 correspond to computing the fitness score and checking it with the threshold, $\gamma$.

\subsection{Correspondences Estimation}
\label{sec:corr_est}
\begin{figure}[!t]
\centering
\vspace{-5pt}
\includegraphics[width=0.85\linewidth]{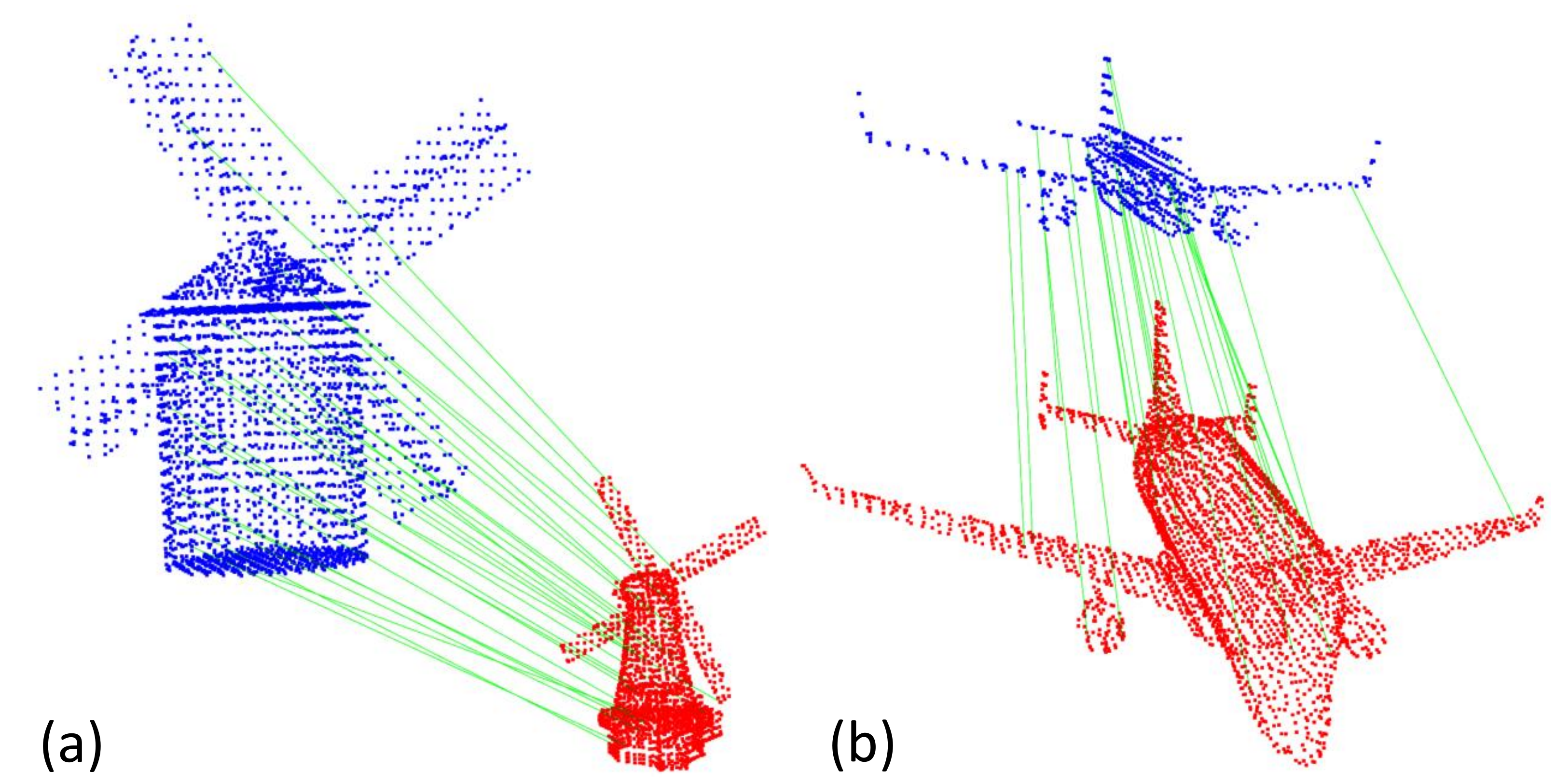}    
\caption{The correspondences (\textcolor{green}{green} lines) estimated between the demonstration model \textcolor{blue}{(blue)} and a target model \textcolor{red}{(red)} of a) Two models of wind turbines; and b) Airbus A320 and Airbus Beluga. For clarity, only a subset of the correspondences has been visualized; 33 out of the 1120 correspondences between the two wind turbine models, and 24 out of the 693 correspondences between the airplane models have been shown.}  
\label{fig:corr_est}
\vspace{-7mm}
\end{figure}

Once the models are found to have similarities using ICP, the correspondences between the two models are estimated. When the two models have a low fitness score, it indicates that they align well with each other. The transformation estimated during the ICP from one-point cloud to the other is used to align and superimpose the two-point clouds. Once the two-point clouds are aligned, the closest point in $\kappa_T$ for every point in $\kappa_D$ is computed as the correspondence,  $\mbox{C}_{DT}$ ($\kappa_D\Longleftrightarrow\kappa_T$) between the point clouds. The flow of computing correspondences between the two-point clouds of the structures is described in lines 7 to 12 in Algorithm \ref{alg:corr_alg}. In Fig. \ref{fig:corr_est}, we show the correspondences estimated between the two wind turbines, and models of Airbus A320 \textcolor{blue}{(blue)} airplane and an Airbus Beluga \textcolor{red}{(red)}. Despite the two airplane models having different sizes ($37.6$ $m$ vs $63.10$ $m$ in length), wing span ($34.10$ $m$ vs $60.3$ $m$), and other characteristics (geometry of the fuselages is different), ICP converged due to their identical shape, and meaningful correspondences were found.

\subsection{Target Inspection Trajectory Planning}
\label{sec:traj_comp}
The trajectory planning process involves estimating an inspection trajectory for the target model, after it has been confirmed that the demonstration and target models are similar through the convergence check and their correspondences have been estimated. The pseudo-code for the target inspection planning is presented in Algorithm \ref{alg:targ_alg}.
\begin{algorithm}
\caption{Convergence Check and Correspondence Estimation}
\begin{algorithmic}[1]
\Require Demonstration Model, $\mathbb{M}_D$; Target Model, $\mathbb{M}_T$; Point Cloud of Demonstration Model, $\Psi_D$; and Point Cloud of Target Model, $\Psi_T$
\Ensure Correspondences from $\mathbb{M}_D$ to $\mathbb{M}_T$, $\mbox{C}_{DT}$ 
\State $\kappa_{D}\gets Voxel\; Grid\;Filter(\Psi_D)$\tikzmark{top}
\State $\kappa_{T}\gets Voxel\; Grid\; Filter(\Psi_T)$ 
\State $\alpha\gets scale\; between\; \kappa_{D}\; and\; \kappa_{T}$ 
\State $\kappa_{T}^{scaled}\gets \alpha*\kappa_{T}$
\State $F \gets ICP\; Fitness\; Score(\kappa_{D}, \kappa_{T}^{scaled})$  \tikzmark{bottom}
\If{$F < \gamma$} 
\State $T_{\kappa_{D} \kappa_{T}^{scaled}} \gets ICP\; Transform(\kappa_{D}, \kappa_{T}^{scaled})$ \tikzmark{top1}
\State $\kappa_{T}^{aligned} \gets Transformation(\kappa_{T}^{scaled}, T_{\kappa_{D} \kappa_{T}^{scaled}})$ \tikzmark{right1}\tikzmark{right}
\State $\mbox{C}_{DT} \gets [\;]$
\ForAll {$k \in \mathcal \;\kappa_{D} $}
    \State $p \gets Find\; Closest\; Point(\kappa_{T}^{aligned},k)$
    \State $\mbox{C}_{DT}.append(k,p)$ \tikzmark{bottom1}
\EndFor
\EndIf
\State \textbf{end}
\end{algorithmic}
\label{alg:corr_alg}
\AddNote{top}{bottom}{right}{Convergence Check}
\AddNote{top1}{bottom1}{right1}{Correspondences Estimation}
\vspace{-4mm}
\end{algorithm}

\noindent\textbf{Visibility Estimation for Demonstration Trajectory.~}
The demonstration trajectory, $\mathbb{P}_D$, given as input can inspect either complete or some parts of the demonstration model,  $\mathbb{M}_D$. Therefore, it is necessary to estimate the regions viewed by the trajectory on the surface of the model using visibility. The visibility from a point can be estimated using Frustum Culling based on sensor information such as Field Of View (FOV) and range limitations. For a given point $p \in \mathbb{P}_D$,  the list of points in  $\kappa_D$  that are visible, $v_{D}^{p}$, is estimated. This provides the regions on the surface of the model that is visible from each and every point on the demonstration trajectory. Based on the visibility from every point in the trajectory, the total visibility is $V_{D}$ was computed. In Algorithm \ref{alg:targ_alg}, lines 1 to 3 correspond to estimating the visibility of the demonstration trajectory. 

\noindent\textbf{Demonstration Inspection Viewpoints Estimation.~}
The main objective of estimating demonstration inspection viewpoints, $\mathbb{IV}_D$ is to generate a set of discrete viewpoints around the demonstration model whose total visibility is the same as $V_{D}$. To achieve this, first, the trajectory $\mathbb{P}_D$ is segmented such that points in each segment, view a common region of the model. This is done so that these segments can be approximated into inspection viewpoints later on. 

To estimate, at which points the trajectory is segmented, the visibility from each point is compared with the following points in the trajectory. This starts from the first point in the trajectory and the visibility of this point is compared with the visibility of the subsequent points. The visibility is compared by checking for common points on the surface structure visible from both trajectory points. When the number of common points is less than a threshold, $\lambda$, a segment break is considered at that point. The same process is continued again by comparing this segment break point with its subsequent points until another segment break is found. This continues until the end of the trajectory is reached. In the implementation, the value of $\lambda$ is set such that new segments are formed when the UAVs move and look at new regions of the surface. The threshold $\lambda$ decides what percentage of the new surface is seen as a new segment is considered. 

Once the segments are identified, they are approximated into demonstration inspection viewpoints $\mathbb{IV}_D$ by computing the centroid of all the points in that segment. Lines 4 to 9 of the Algorithm \ref{alg:targ_alg} expound the demonstration inspection viewpoint estimation process. 

\begin{algorithm}
\caption{Target Inspection Trajectory Planning}
\begin{algorithmic}[1]
\Require same as Algorithm 1
\Ensure Target Inspection Trajectory, $\mathbb{P}_T$ 
\ForAll {$p\in \mathcal \; \mathbb{P}_{D} $} \tikzmark{top2}
    \State $v_{D}^{p} \gets visibility(\kappa_{D},p)$
\EndFor
\State $V_{D} \gets  \bigcup_{i=1}^{n}v_{D}^{i}$\tikzmark{bottom2}
\State $seg_{start} \gets v_{D}^{1}$\tikzmark{top3}
\State $segments \gets [\;]$
\ForAll {$q\in \mathcal \; \mathbb{P}_{D} $}
    \If{$n(seg_{start} \cap v_{D}^{q}) \cap n(V_{D})  < \lambda$} \tikzmark{right2}\tikzmark{right3}\tikzmark{right4}
        \State $segments.append(seg_{start}, v_{D}^{q})$
        \State $seg_{start} \gets v_{D}^{q}$
    \EndIf
\EndFor
\State $\mathbb{IV}_D \gets centroid(segments)$\tikzmark{bottom3}
\State $\mu_D, \sigma_D \gets Distribution\; of\; \kappa_D$\tikzmark{top4}
\State $\mu_T, \sigma_T \gets Distribution\; of\; \kappa_T$
\State $\mathbb{IV}_D \gets (\mathbb{IV}_D - \mu_{D}) / {\sigma_{D}} $
\State $\mathbb{IV}_T \gets z\mathbb{IV}_D * \sigma_{T} + \mu_{T}$\tikzmark{bottom4}
\State $\mathbb{P}_T \gets optimize(\mathbb{IV}_T)$
\State \textbf{end}
\end{algorithmic}
\label{alg:targ_alg}
\AddNot{top2}{bottom2}{right2}{Visibility Estimation}
\AddNot{top3}{bottom3}{right3}{Demonstration Inspection Viewpoints}
\AddNot{top4}{bottom4}{right4}{Target Inspection Viewpoints}
\vspace{-4mm}
\end{algorithm}

\noindent\textbf{Target Inspection Viewpoints Estimation.~}
Once we obtain the Demonstration Inspection Viewpoints $\mathbb{IV}_D$, the next goal is to estimate the Target Inspection Viewpoints $\mathbb{IV}_T$ using $\mathbb{IV}_D$. $\mathbb{M}_D$ and $\mathbb{M}_T$ are structures that are very similar, and hence each point of $\mathbb{IV}_T$ ideally views a region of $\kappa_T$ that is analogous to the regions viewed in $\kappa_D$ at every point in $\mathbb{IV}_D$. Since $\mathbb{M}_D$ and $\mathbb{M}_T$ vary in scales and geometries; feature scaling is used to standardize ($z$-score normalize) both the point clouds $\kappa_D$ and $\kappa_T$ to the same definite scale ranges. This allows us to analyze standardized values of $\mathbb{IV}_D$ at a scale that would fit for both $\kappa_D$ and $\kappa_T$. 

Let $\kappa_D$ be distributed with mean $\mu_D$ and standard deviation $\sigma_D$, and $\kappa_T$ be distributed with mean $\mu_T$ and standard deviation $\sigma_T$. The Demonstration Inspection Viewpoints $\mathbb{IV}_D$ are transformed to standard score $z$$\mathbb{IV}_D$ using,  
\begin{equation}
   z\mathbb{IV}_D = \dfrac{\mathbb{IV}_D - \mu_{D}} {\sigma_{D}}. 
\end{equation}

Since feature scaling is performed between $\kappa_D$ and $\kappa_T$, we can estimate $\mathbb{IV}_T$ by scaling back the standard score $z$$\mathbb{IV}_D$ to a distribution with mean $\mu_T$ and standard deviation $\sigma_T$ as, 
\begin{equation}
   \mathbb{IV}_T = z\mathbb{IV}_D \cdot \sigma_{T} + \mu_{T}.
\end{equation}

This is akin to inverse transforming $z$$\mathbb{IV}_D$ with respect to the distribution of $\kappa_T$. Lines 4 to 9 of Algorithm \ref{alg:targ_alg} delineate the process of estimating target inspection viewpoints.

\noindent\textbf{Optimizing Target Inspection Viewpoints.~}
The target inspection viewpoints generated should replicate the overall structure of the demonstration path and gives a coarse path for inspection. But, this may not ensure that each target inspection viewpoint views a similar region to the demonstration inspection viewpoint using which it was generated. This is because the inspection viewpoints are translated from one structure to another based on the distribution of the entire structure and not the local distribution near that inspection viewpoint. Therefore, the positions of the target inspection viewpoints are further refined using the Gauss-Newton algorithm.

Let $iv_{T}^{j}$ be a viewpoint in the target inspection viewpoint generated based on a demonstration inspection viewpoint $iv_{D}^{j}$ with visibility $v_D^j$. Now, the goal of this optimization step is to refine $iv_{T}^{j}$ such that it can view the corresponding points of $v_D^j$ in the target model ($v_T^j$) as per the correspondences, $C_{DT}$. $v_D^j$ and $v_T^j$ are standardized using z-score normalization (as elaborated previously). The distances to each and every point in $v_D^j$ and $v_T^j$ from $iv_{D}^{j}$ and $iv_{T}^{j}$ respectively, are defined as $d_{D}^j$ and $d_{T}^j$.
These distances are used as the residual $r_j$ for the Gauss-Newton,
\begin{equation}
    r_{j} = d_{T}^j - d_{D}^j.
\end{equation}

With $r_j$, as the residual the position of $iv_{T}^{j}$ is refined iteratively. This makes $iv_{T}^{j}$ have a similar view to that of $iv_{D}^{j}$ and ultimately have a similar overall inspection to that of the demonstration. 

Finally, once the position of the target inspection viewpoints is optimized, this sequence of viewpoints is used to construct the trajectory. The time for each point on the trajectory is computed such that UAV resembles a similar velocity with which the UAV was teleoperated between two demonstration inspection viewpoints. 

\section{Experiments and Results}
\label{sec:expt}
We validate the proposed framework using a range of 3D models of large structures commonly inspected, including airplanes, wind turbines, and ships in a simulated environment. Additionally, we include representative models of a cubical building and a bridge in a real-world indoor environment. We selected these models to represent structures with different geometric complexities. All the 3D models used in the simulated environment were accurately scaled to reflect their real-world counterparts. 

To evaluate our proposed method, we compare it against a simple scaling-based baseline method, where the demonstration trajectory is scaled non-uniformly using the ratio of the sizes of the bounding boxes of the demonstration and target models. We evaluate both methods by measuring the percentage of the points on the surface of the structure viewed by the demonstration and target trajectories, using the correspondence information between the points. A value of $100\%$ indicates that every point on the surface viewed by the demonstration trajectory was also viewed by the target trajectory. 

We also measure the Fréchet distance \cite{alt1995computing} between demonstrated and target inspection trajectories. The Fréchet distance measures the similarity between the shapes of trajectories, rather than just the location of the points in them. In our measurements, we standardize ($z$-score normalize) both the trajectories and then measure the Fréchet distance. The resulting distance is a normalized measure of the similarity between the two trajectories. A Fréchet distance of $0$ indicates that both trajectories have the same shape. We exclude the baseline method from this comparison because it scales the demonstrated trajectory and always returns a Fréchet distance of $0$. 

\subsection{Simulation Setup}
The experiment setup was implemented using Webots. It was used to obtain demonstration trajectories through teleoperation and also to validate the generated Target Path. An actual human operator teleoperated the UAV and generated demonstration trajectories. The parameters for the visibility computation: the horizontal and vertical FOV were set to $75$°; safety distance to $2$ $m$; and maximum viewing distance to $50$ $m$. The parameters used are similar to those in the literature \cite{jing2020multi}. 

\subsection{Real-world Experimental Setup}
\begin{figure}[t]
\centering
    \begin{subfigure}{0.80\linewidth} 
        \includegraphics[width=\linewidth]{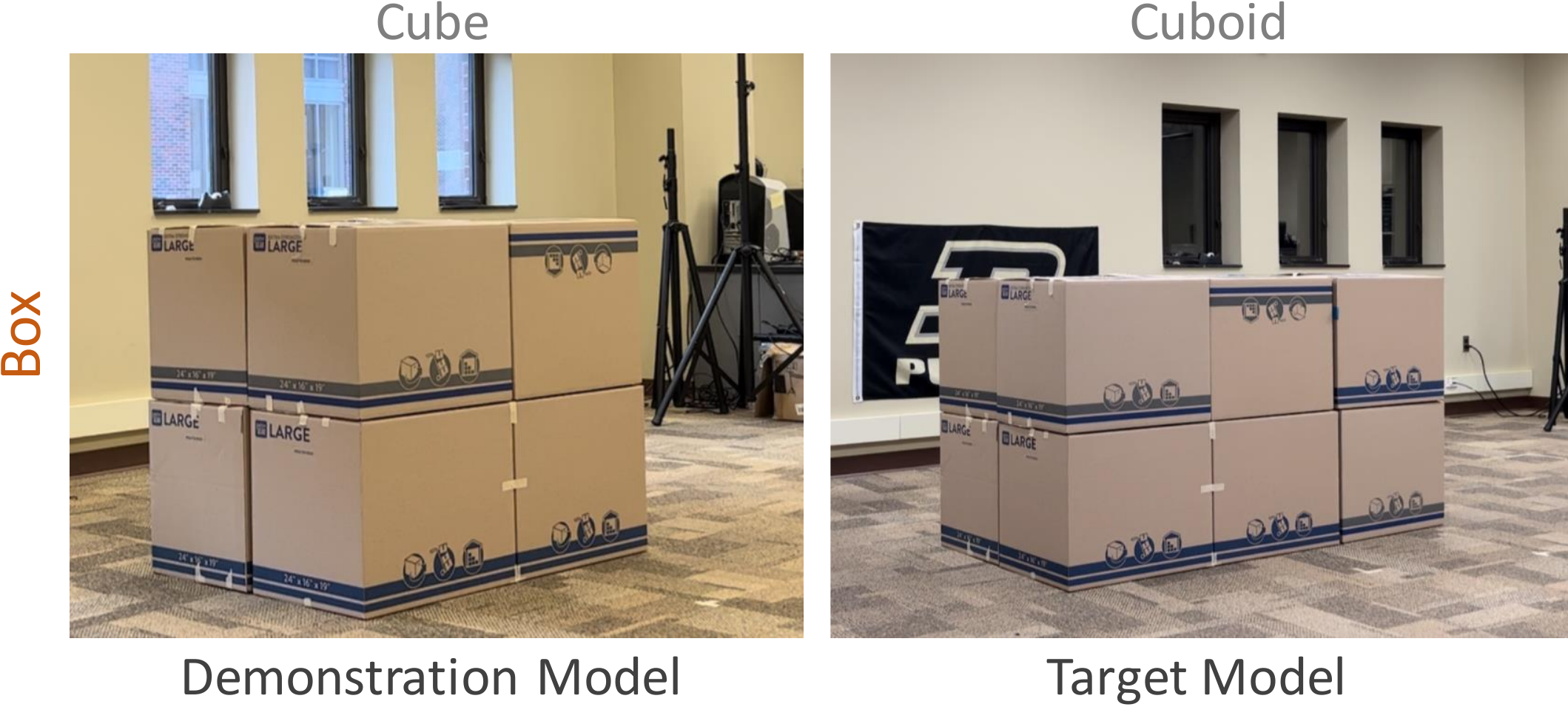}
            \vspace{-6mm}
        \caption{}
        \label{fig:box}
    \end{subfigure}

    \begin{subfigure}{0.80\linewidth} 
        \includegraphics[width=\linewidth]{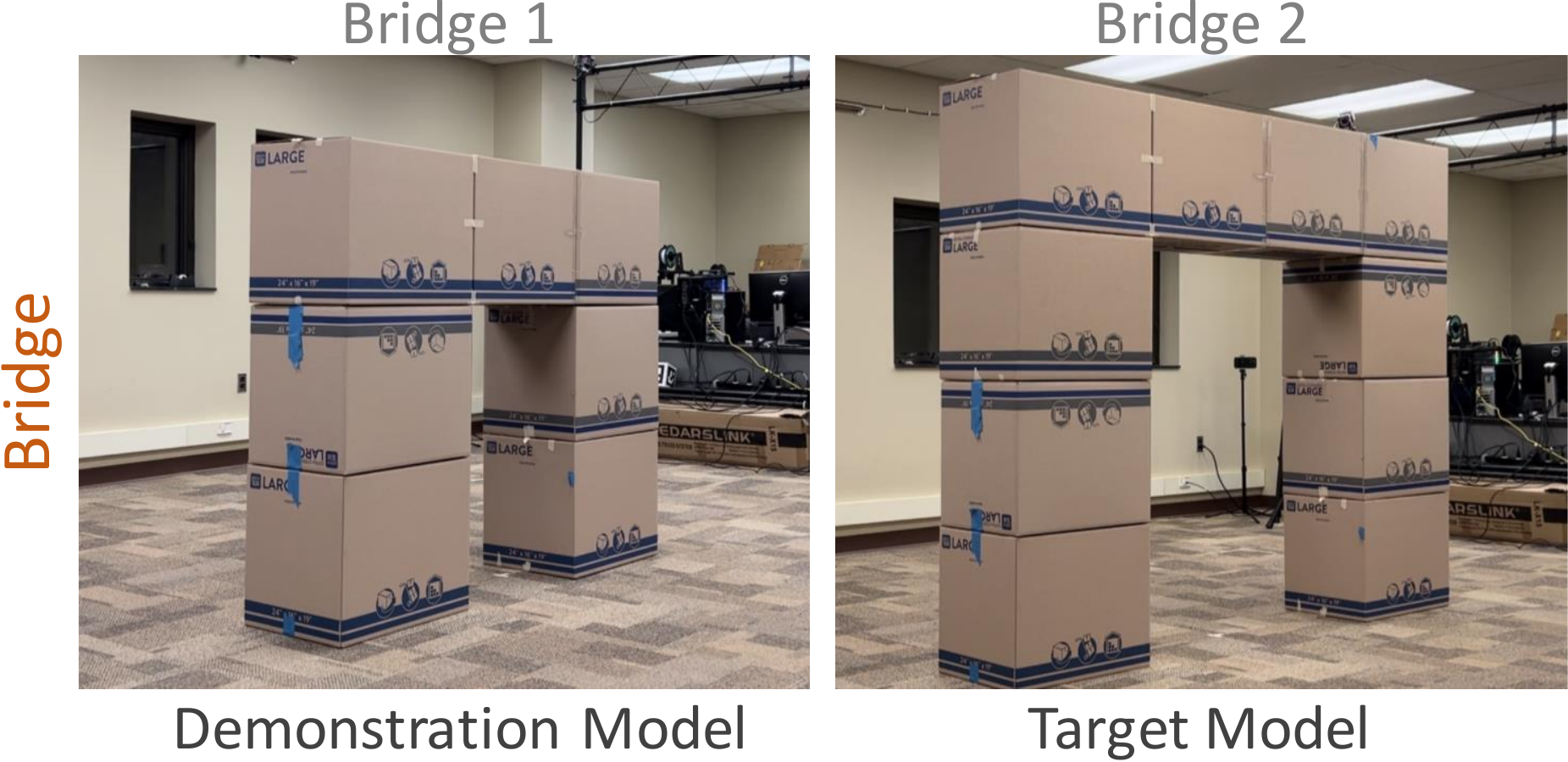}
            \vspace{-6mm}
        \caption{}
        \label{fig:bridge}
    \end{subfigure}
\caption{a) box and b) bridge structures used for real-world experiments.}
\label{fig:expt_setup}  
\vspace{-6mm}
\end{figure}

Real-world experiments were conducted in an indoor environment using the Parrot Bebop 2 UAV platform. The UAV's localization was achieved through Vicon motion capture, and the Robot Operating System (ROS) framework was employed for UAV control. To enhance realism, structures resembling a cubical building and a bridge were constructed using cardboard boxes and utilized for the experiments. Fig. \ref{fig:expt_setup} shows the structures used for the experiments. 

\begin{figure*}[t]
\centering
\includegraphics[width=\linewidth]{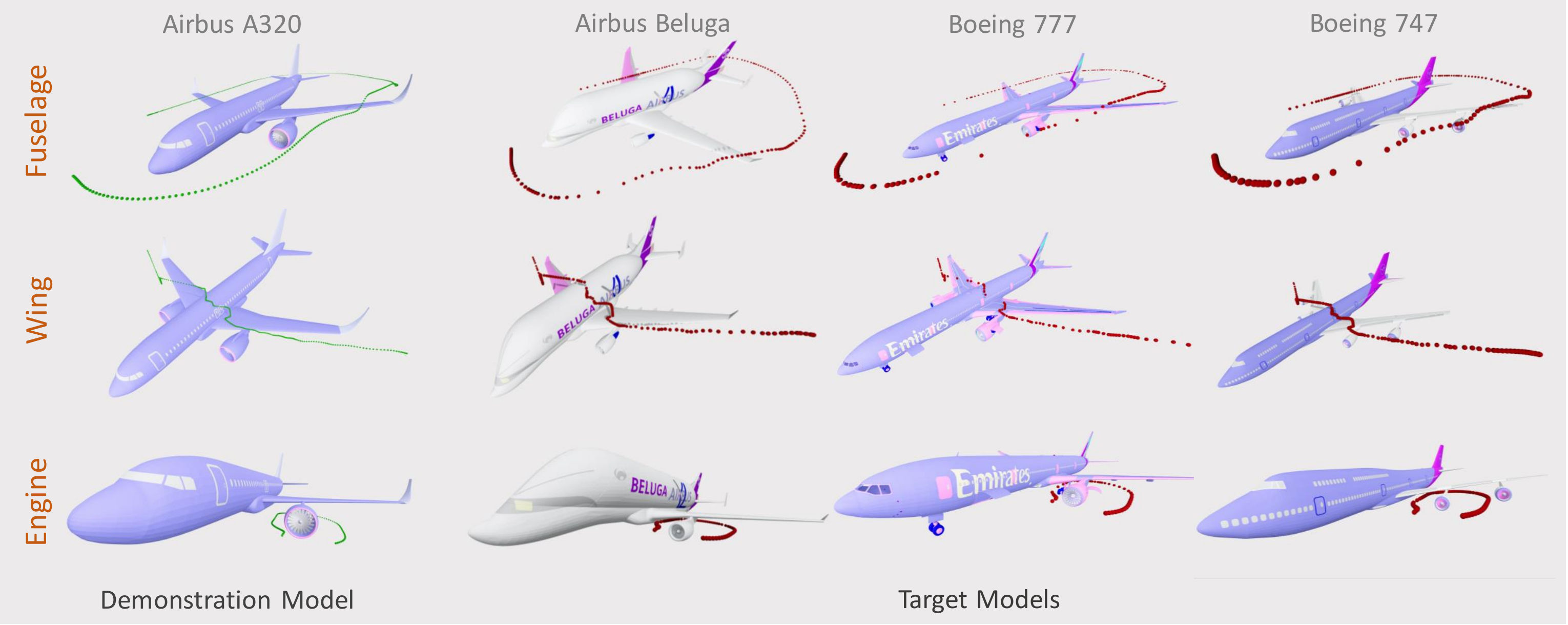}  
\caption{Airplane Inspection: Demonstration trajectory (\textcolor{green}{green}) for inspecting an Airbus A320 airplane (left most) and the generated target trajectories (\textcolor{Maroon}{red}) for inspecting an Airbus Beluga, a Boeing 777 and a Boeing 747 airplanes (second left to right), focusing the fuselage, the wing and the engine of the respective airplanes.}  
\label{fig:airplane}
\vspace{-5mm}
\end{figure*}

\subsection{Experimental Results}
 \noindent\textbf{Airplane.~}
In this experiment, we used four different airplane models: Airbus A320, Airbus Beluga, Boeing 777, and Boeing 747. These models of airplanes were chosen since they exhibit differences in their geometry and scale. The Airbus A320 model was used for the demonstration and similar trajectories were generated for the other three models. Three types of trajectories inspecting the fuselage, wing, and engine of the airplane were used as the demonstrations. Fig. \ref{fig:airplane} shows the demonstration model of Airbus A320 and the three demonstration trajectories along with the target trajectory generated for the other three models: Airbus Beluga, Boeing 777, and Boeing 747. Based on different parts of the airplane that are commonly inspected, trajectories inspecting different parts of the airplane were used as the demonstration.   

First, we demonstrated a trajectory to move around the airplane and inspect the fuselage of the airplane. The planner was able to generate successful trajectories for all three target models and scaled them in accordance with the scale of the airplane as shown in the fuselage inspection first row of Fig. \ref{fig:airplane}. The proposed methods achieved over $98\%$ coverage of the demonstrated trajectory, despite the differences in the geometry of the airplanes.

Second, a trajectory to inspect the wing of an airplane was demonstrated. In the demonstration, the trajectory started from the far end of the left-wing, proceeded to the center, and then moved parallel to the right wing. The trajectory got too close to the center of the airplane while transitioning from the left side of the airplane to the right. Simply translating the normalized viewpoints from the demonstration to the target or scaling the demonstrated trajectory might result in a collision with the surface of the airplane. In contrast, our framework generated trajectories that avoid collision with the surface, because the optimization part of the framework adjusts the viewpoints to maintain a certain distance from the surface based on visibility from that viewpoint and this handles potential collisions. The demonstration and target trajectories are shown in the wing inspection, the second row of Fig. \ref{fig:airplane}.

Finally, a trajectory was demonstrated to inspect the left engine of an airplane. Despite the variations in the target model, the planner was able to generate relevant trajectories to inspect the engines of the airplanes. The demonstrated and target trajectories are shown in the engine inspection of Fig. \ref{fig:airplane}. Inspecting the engine was an interesting case since all the airplane models have left and right engines. While generating the target trajectory, there was not any confusion between the left and the right engines since the correspondences were found after aligning the models with one another. Further, Boeing B747 (bottom row rightmost in Fig. \ref{fig:airplane}) is a Twin-Engine jet and has two engines on each side. When the models were aligned while computing the correspondences, the engine from the demonstration model of A320 was closer to the inner engine and hence, the inspection trajectory was generated for the inner engine and not the outer one.

\begin{figure}[t]
\centering
\vspace{2mm}
    \begin{subfigure}{0.875\linewidth} 
        \includegraphics[width=\linewidth]{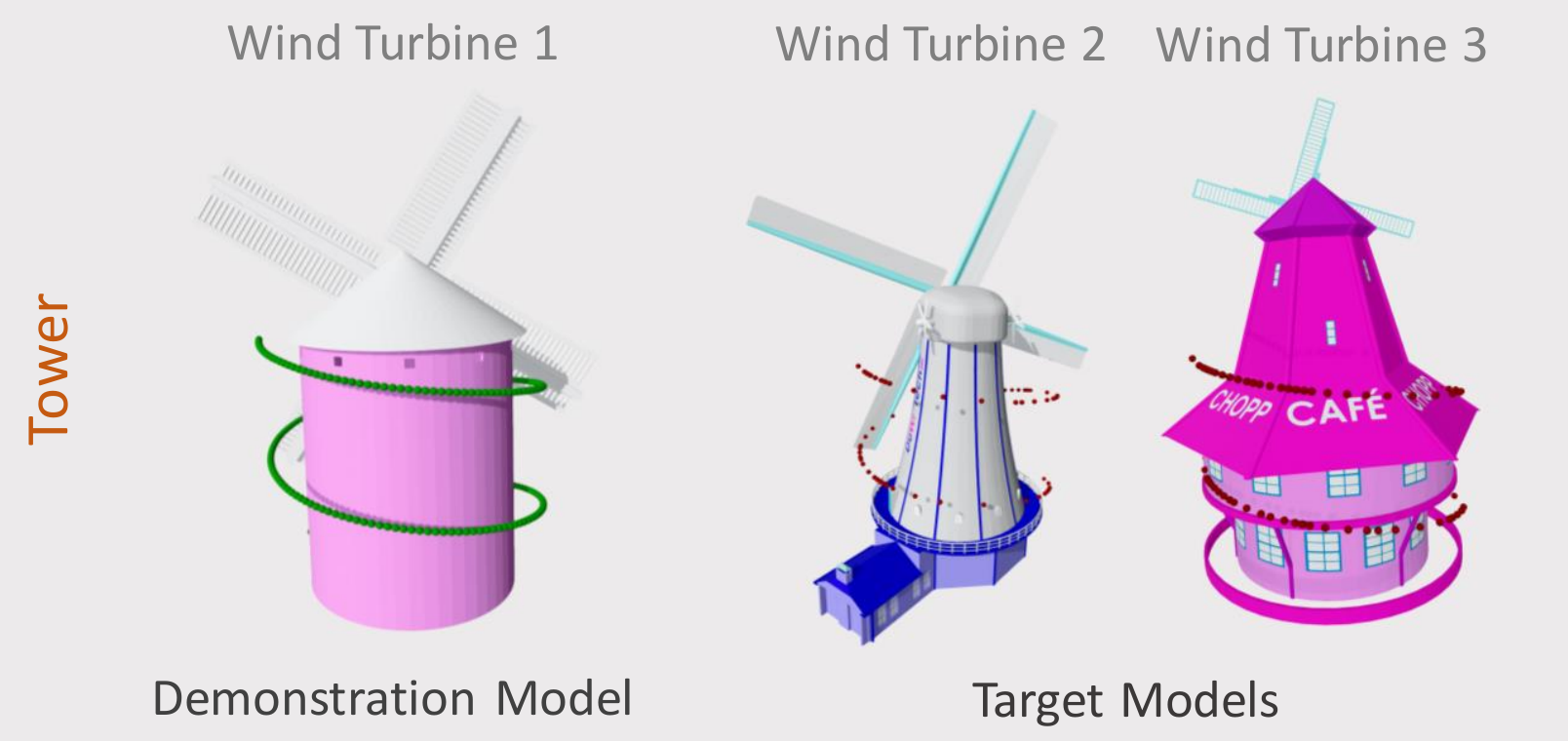}
        \vspace{-6mm}
        \caption{}
        \label{fig:wind}
    \end{subfigure}
    \begin{subfigure}{0.975\linewidth} 
        \includegraphics[width=\linewidth]{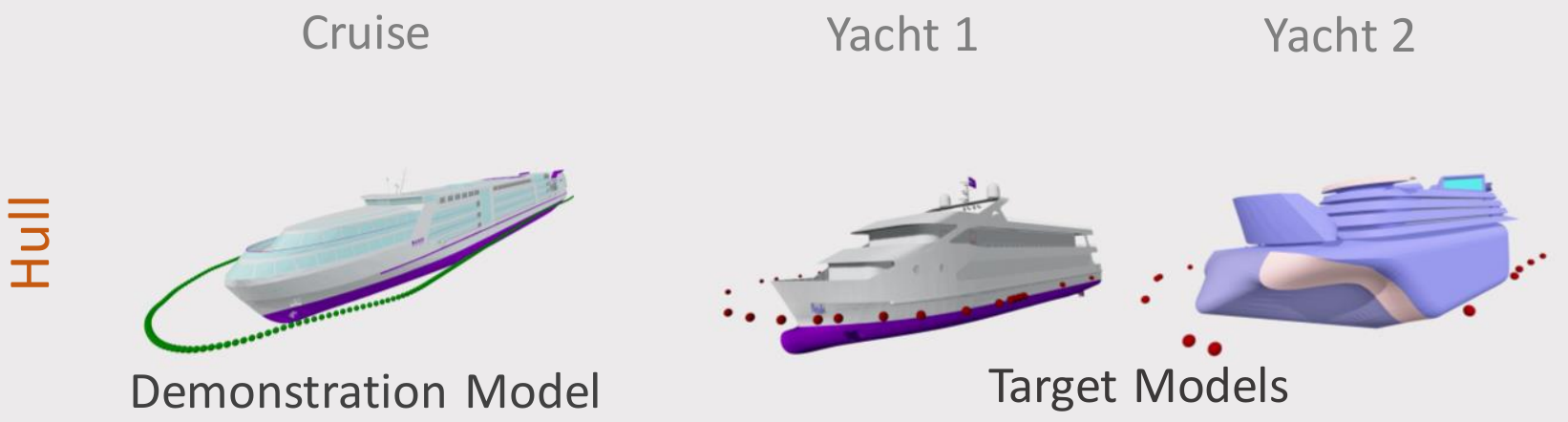}
        \vspace{-6mm}
        \caption{}
        \label{fig:ship}
    \end{subfigure}
\caption{a) Wind Turbine and b) Ship Inspections along with demonstration trajectory (\textcolor{green}{green}) and the generated target trajectories (\textcolor{Maroon}{red}).}
\label{fig:wind_ship}  
\vspace{-3mm}
\end{figure}

\noindent\textbf{Wind Turbine.~}
The framework was used to generate inspection paths for the inspection of wind turbine towers. A spiral path around the tower of a wind turbine (wind turbine 1) was demonstrated and paths were generated for two other wind turbines with different structures (wind turbine 2 \& 3). Despite these differences in the structure as shown in \ref{fig:wind}, the cylindrical tower part of the wind turbines had close correspondence and the demonstration trajectory was replicated to inspect the tower regions of the other wind turbines too. 

\noindent\textbf{Ship.~}
Inspection of the hull of a ship was also conducted. A demonstration was made on a cruise ship and new paths were generated for inspecting the hull of yachts. The demonstration path from a long cruise with numerous waypoints was transferred into a concise path with fewer waypoints as seen in Fig. \ref{fig:ship}. This is because we use high-level inspection points that give the same visibility as the demonstrated path and not each and every point in the demonstrated path.  

\begin{figure}[t]
\centering
    \begin{subfigure}{0.92\linewidth} 
        \includegraphics[width=\linewidth]{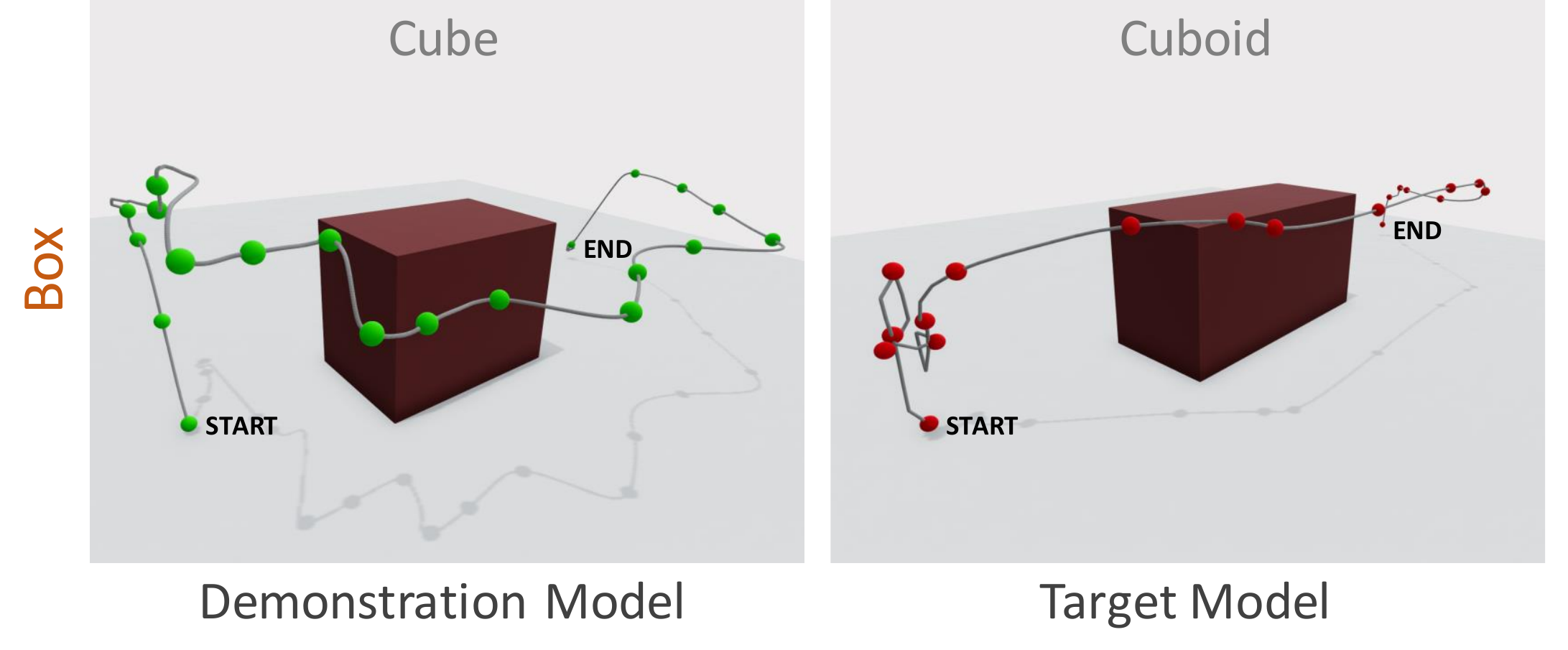}
        \vspace{-6mm}
        \caption{}
        \label{fig:box}
    \end{subfigure}
    \begin{subfigure}{0.90\linewidth} 
        \includegraphics[width=\linewidth]{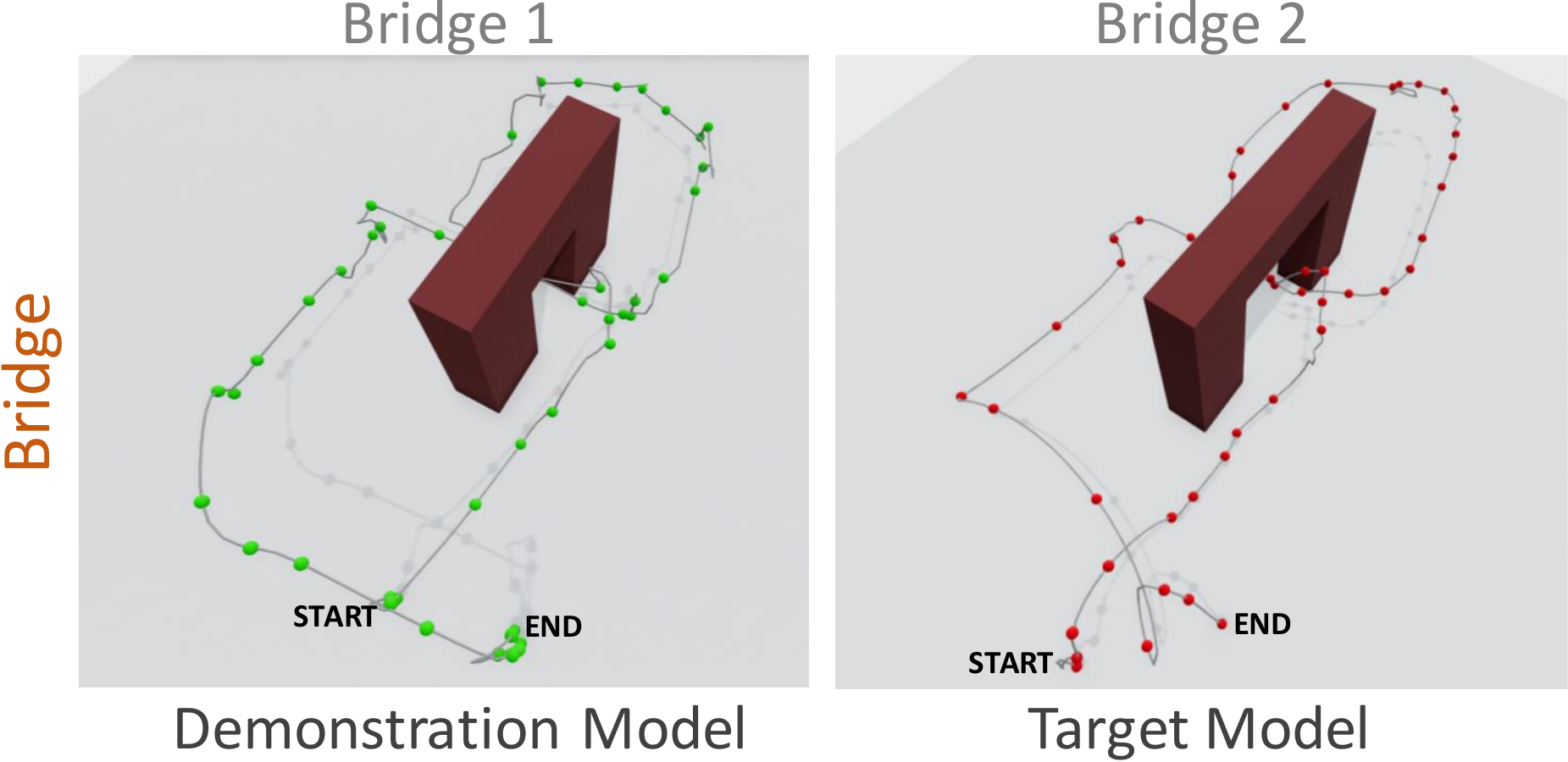}
       \vspace{-6mm}
        \caption{}
        \label{fig:bridge}
    \end{subfigure}
\caption{a) box and b) bridge structures inspection along with demonstration trajectory (\textcolor{green}{green}) and the generated target trajectories (\textcolor{Maroon}{red}) from the real-world experiments.}
\label{fig:real_expt}  
\vspace{-4mm}
\end{figure}

\begin{table}[h]
\caption{Percentage of the surface points that were covered by the target trajectory using our framework and the baseline.}
\label{tab:cover_percentage}
\begin{tabular}{cccccc}
Object                                                                   & \begin{tabular}[c]{@{}c@{}}Inspection \\ Type\end{tabular} & \begin{tabular}[c]{@{}c@{}}Demonstrated\\ Model\end{tabular}               & \begin{tabular}[c]{@{}c@{}}Target\\ Model\end{tabular}   & Ours  & \begin{tabular}[c]{@{}c@{}}Base-\\ line\end{tabular} \\ \hline\hline
\multirow{17}{*}{Plane}                                                   & \multirow{5}{*}{Fuselage}                                    &                                                                            & \begin{tabular}[c]{@{}c@{}}Airbus \\ Beluga\end{tabular} & 98.3  & 88.8                                                 \\
                                                                         &                                                            & Airbus A320                                                                & \begin{tabular}[c]{@{}c@{}}Boeing \\ 747\end{tabular}    & 99.8  & 98.8                                                 \\
                                                                         &                                                            &                                                                            & \begin{tabular}[c]{@{}c@{}}Boeing \\ 777\end{tabular}    & 98.3  & 72.8                                                 \\ \cline{2-6} 
                                                                         & \multirow{5}{*}{Wings}                                     &                                                                            & \begin{tabular}[c]{@{}c@{}}Airbus \\ Beluga\end{tabular} & 96.61 & 86.6                                                 \\
                                                                         &                                                            & Airbus A320                                                                & \begin{tabular}[c]{@{}c@{}}Boeing \\ 747\end{tabular}    & 99.5  & 99.2                                                 \\
                                                                         &                                                            &                                                                            & \begin{tabular}[c]{@{}c@{}}Boeing \\ 777\end{tabular}    & 94.3  & 78.9                                                 \\ \cline{2-6} 
                                                                         & \multirow{5}{*}{Engine}                                      &                                                                            & \begin{tabular}[c]{@{}c@{}}Airbus \\ Beluga\end{tabular} & 92.8  & 83.6                                                 \\
                                                                         &                                                            & Airbus A320                                                                & \begin{tabular}[c]{@{}c@{}}Boeing \\ 747\end{tabular}    & 99.9  & 98.6                                                 \\
                                                                         &                                                            &                                                                            & \begin{tabular}[c]{@{}c@{}}Boeing \\ 777\end{tabular}    & 92.8  & 83.6                                                 \\ \hline
\multirow{3}{*}{\begin{tabular}[c]{@{}c@{}}Wind \\ Turbine\end{tabular}} & \multirow{2}{*}{Tower}                                     & \multirow{3}{*}{\begin{tabular}[c]{@{}c@{}}Wind \\ Turbine 1\end{tabular}} & \begin{tabular}[c]{@{}c@{}}Wind\\ Turbine 2\end{tabular} & 99.7  & 98.7                                                 \\
                                                                         &                                                            &                                                                            & \begin{tabular}[c]{@{}c@{}}Wind\\ Turbine 3\end{tabular} & 100   & 98                                                   \\ \hline
\multirow{2}{*}{Ship}                                                    & \multirow{2}{*}{Hull}                                      & \multirow{2}{*}{Cruise}                                                    & Yacht 1                                                  & 99.5  &      99.5                                                \\
                                                                         &                                                            &                                                                            & Yacht 2                                                  & 100  & 100                                                     \\ 
\hline
\multirow{1}{*}{Box}                                                    & \multirow{1}{*}{Sides}                                      & \multirow{1}{*}{Cube}                                                    & Cuboid                                                  & 100  &      100                                                
                                                                                                                             \\ 
\hline
\multirow{1}{*}{Bridge}                                                    & \multirow{1}{*}{Piles}                                      & \multirow{1}{*}{Bridge 1}                                                    & Bridge 2                                                  & 99.8  &      98.4                                                
                                                                                                                             \\\hline\hline
\end{tabular}
\vspace{-2mm}
\end{table}

\noindent\textbf{Real-world Indoor Experiment.~}
In the real-world indoor experiments, a box-like structure resembling a simple building and a bridge-like structure were utilized. For the box structure, the inspection demonstration involved navigating a cube structure to generate paths for inspecting a cuboid structure. An inspection path in the shape of ``U" was demonstrated to inspect the sides of the cubical structure, and a target path was generated for the cuboid, as illustrated in Fig. \ref{fig:box}. Regarding the bridge structure inspection, the goal was to inspect the piles of two variations of the bridge-like structure. A loop in the shape of the number ``8" was employed to inspect all sides of the piles, as depicted in Fig. \ref{fig:bridge}. In both cases, the demonstration trajectories were characterized by noise and lack of smoothness. However, the generated target trajectories exhibited smoother trajectories compared to the demonstration trajectories. This improvement can be attributed to the selection of inspection viewpoints, which helps in filtering out noisy and outlier points in the trajectory. 

In Table \ref{tab:cover_percentage} and Table \ref{tab:frechet}, we summarize the percentage of surface points that were covered and the Fréchet distance, for target trajectories generated using our method and the baseline for various inspection models described above. The readers are recommended to refer to the supplementary video of the paper for human demonstration of the paths and UAVs inspecting using the target paths.

\begin{table}[h]
\centering
\caption{Fréchet Distance between demonstrated inspection trajectory and target inspection trajectory is shown for demonstration models w.r.t multiple target models (Fréchet Distance is the normalized measure of the similarity between the two trajectories). }
\label{tab:frechet}
\begin{tabular}{ccccc}
Object                                                                   & \begin{tabular}[c]{@{}c@{}}Inspection \\ Type\end{tabular} & \begin{tabular}[c]{@{}c@{}}Demonstrated\\ Model\end{tabular}               & \begin{tabular}[c]{@{}c@{}}Target\\ Model\end{tabular}   & \begin{tabular}[c]{@{}c@{}}Fréchet \\ distance\end{tabular} \\ \hline\hline
\multirow{17}{*}{Plane}                                                   & \multirow{5}{*}{Fuselage}                                    &                                                                            & \begin{tabular}[c]{@{}c@{}}Airbus \\ Beluga\end{tabular} & 0.7                                                         \\
                                                                         &                                                            & Airbus A320                                                                & \begin{tabular}[c]{@{}c@{}}Boeing \\ 747\end{tabular}    & 1.09                                                        \\
                                                                         &                                                            &                                                                            & \begin{tabular}[c]{@{}c@{}}Boeing \\ 777\end{tabular}    & 0.75                                                        \\ \cline{2-5} 
                                                                         & \multirow{5}{*}{Wings}                                     &                                                                            & \begin{tabular}[c]{@{}c@{}}Airbus \\ Beluga\end{tabular} & 0.74                                                        \\
                                                                         &                                                            & Airbus A320                                                                & \begin{tabular}[c]{@{}c@{}}Boeing \\ 747\end{tabular}    & 0.58                                                        \\
                                                                         &                                                            &                                                                            & \begin{tabular}[c]{@{}c@{}}Boeing \\ 777\end{tabular}    & 0.74                                                        \\ \cline{2-5} 
                                                                         & \multirow{5}{*}{Engine}                                      &                                                                            & \begin{tabular}[c]{@{}c@{}}Airbus \\ Beluga\end{tabular} & 1.45                                                        \\
                                                                         &                                                            & Airbus A320                                                                & \begin{tabular}[c]{@{}c@{}}Boeing \\ 747\end{tabular}    & 1.9                                                         \\
                                                                         &                                                            &                                                                            & \begin{tabular}[c]{@{}c@{}}Boeing \\ 777\end{tabular}    & 1.47                                                        \\ \hline
\multirow{3}{*}{\begin{tabular}[c]{@{}c@{}}Wind \\ Turbine\end{tabular}} & \multirow{2}{*}{Tower}                                     & \multirow{3}{*}{\begin{tabular}[c]{@{}c@{}}Wind \\ Turbine 1\end{tabular}} & \begin{tabular}[c]{@{}c@{}}Wind\\ Turbine 2\end{tabular} & 0.63                                                        \\
                                                                         &                                                            &                                                                            & \begin{tabular}[c]{@{}c@{}}Wind\\ Turbine 3\end{tabular} & 0.75                                                        \\ \hline
\multirow{2}{*}{Ship}                                                    & \multirow{2}{*}{Hull}                                      & \multirow{2}{*}{Cruise}                                                    & Yacht 1                                                  & 1.85                                                        \\
                                                                         &                                                            &                                                                            & Yacht 2                                                  &        1.05 
\\ \hline
\multirow{1}{*}{Box}                                                    & \multirow{1}{*}{Sides}                                      & \multirow{1}{*}{Cube}                                                    & Cuboid                                                  & 0.54                                                        \\ \hline 
\multirow{1}{*}{Bridge}                                                    & \multirow{1}{*}{Piles}                                      & \multirow{1}{*}{Bridge 1}                                                    & Bridge 2                                                  & 0.73                                                        \\
\hline\hline
\end{tabular}
\vspace{-2mm}
\end{table}

\subsection{Discussion}
\label{sec:limi}
While our method is robust in generating trajectories for numerous structures, it is limited to similar structures that can be aligned using ICP.  Without proper alignment, correct correspondences cannot be found, and optimizing the viewpoints may result in poor inspection viewpoint. During experiments, we also found that the path generated may not inspect the desired regions as per the user's expectations when there are multiple similar features in the structure. For example, the engine inspection path generated for the Boeing 747 model in Fig. \ref{fig:airplane} (right-bottom), may not be ideal if the intent was to inspect the outer engine of the airplane. In such scenarios, it is advisable to demonstrate on a twin-engine airplane or other relevant models. 

Although optimizing the inspection viewpoint refines the position of the viewpoints and enhances visibility, it is constrained by the camera's field of view. Regions visible to the UAV based on the camera's viewing range dictate the optimization process's direction. This process does not consider regions not visible to the UAV, which may result in objectionable paths. For instance, if the structure has parts behind the UAV's position, those parts may not be visible, and the path could intersect or collide with them. To overcome such situations, it is wise to use a camera with a wide or $360^o$ visibility.

Furthermore, the proposed framework's applicability can be extended to 2D planning and can be used for ground robots as well. This is demonstrated in the inspection of the ship's hull in \ref{fig:ship}, where the UAVs move almost in a fixed plane both in the demonstration and the target.

\section{Conclusion}
\label{sec:conclusion}
In this paper, we presented UPPLIED, a practically–oriented visual structural inspection path planning framework capable of learning inspection trajectories for complex 3D structures based on expert demonstration. The method was tested in simulated and real-indoor environments with structures that usually require visual inspection. Through these experiments, it was demonstrated that the proposed framework generates paths that inspect similar regions to that of the demonstration trajectory on the target structure. The proposed framework enables the sim-to-real transfer of demonstrations. An expert demonstration in a simulated setup could be transferred to a real-world structure, given that the 3D model of the structure is available. We expect the proposed method to give rise to new innovative directions in the field of structural inspection, and potentially, an expert demonstration trajectory database could be created in the future. Other future work can focus on developing models that can further generalize trajectories and create new trajectories even when the ICP convergence check fails to satisfy them. 
\vspace{-2mm}
\bibliographystyle{IEEEtran}
\bibliography{references}
\end{document}